# Temporal Coherent and Graph Optimized Manifold Ranking for Visual Tracking

**Bo Jiang, Doudou Lin, Bin Luo** and **Jin Tang***
School of Computer Science and Technology, Anhui University, Hefei, China
jiangbo@ahu.edu.cn, ahu_lindd@163.com, luobin@ahu.edu.cn, tj@ahu.edu.cn

## Abstract

Recently, weighted patch representation has been widely studied for alleviating the impact of background information included in bounding box to improve visual tracking results. However, existing weighted patch representation models generally exploit spatial structure information among patches in each frame separately which ignore (1) unary feature of each patch and (2) temporal correlation among patches in different frames. To address this problem, we propose a novel unified temporal coherence and graph optimized ranking model for weighted patch representation in visual tracking problem. There are three main contributions of this paper. First, we propose to employ a flexible graph ranking for patch weight computation which exploits both structure information among patches and unary feature of each patch simultaneously. Second, we propose a new more discriminative ranking model by further considering the temporal correlation among patches in different frames. Third, a neighborhood preserved, low-rank graph is learned and incorporated to build a unified optimized ranking model. Experiments on two benchmark datasets show the benefits of our model.

## 1 Introduction

As an important research topic in computer vision, visual tracking has been extensively studied and many tracking methods have been developed. However, it still remains challenging due to large appearance variations of the target object and disturbance in presence of cluttered background. Existing visual tracking methods usually adopt tracking-by-detection framework which aims to localize the target object via a bounding box by using a classifier during the tracking process. One issue is that the bounding box is difficult to describe the target object accurately due to the irregular shape of the target object and thus usually introduces undesired background information, which usually degrades the effectiveness of the classifier and tracking results.

To address this issue, many methods have been developed to alleviate the impact of background information [Dorin *et al.*, 2003; Hare *et al.*, 2016; He *et al.*, 2013; Yuan *et al.*, 2015; Zhang *et al.*, 2014; Kim *et al.*, 2015; Liu *et al.*, 2017]. One kind of popular methods is to suppress the impact of background by generating robust feature descriptor for bounding box [Dorin *et al.*, 2003; He *et al.*, 2013; Yuan *et al.*, 2015]. Recently, patch based representation has been studied to produce robust object descriptor [Li *et al.*, 2015; He *et al.*, 2017; Lee *et al.*, 2014; Kim *et al.*, 2015; Li *et al.*, 2017a; 2017b]. Among them, one kind of popular methods is to develop weighted patch representation to alleviate the impact of background information and thus improve tracking results [Kim *et al.*, 2015; Li *et al.*, 2017a; 2017b]. These methods generally aim to assign different foreground weights to the patches in bounding box feature representation to alleviate the impact of patches that are belonging to background. For example, Kim et al. [Kim *et al.*, 2015] proposed to construct neighborhood graph to represent the patches of bounding box and employed a random walk model to obtain different weights for the patches. Li et al. [Li *et al.*, 2017a] proposed to learn a low-rank sparse graph for target object representation to obtain more robust weight computation. Li et al. [Li *et al.*, 2017b] provided an improved graph learning model by further considering local and global relationship information among patches together.

However, existing weighted patch representation models have two main limitations. First, they only exploit structure information among patches in weighted patch representation which neglect the unary features of patches explicitly. Second, they generally learn the representation of bounding box in each frame separately which ignores the temporal correlation among different frames. To address these limitations, we propose a novel unified temporal coherence and graph optimized ranking model to obtain robust weighted patch representation for visual tracking. First, we employ a flexible graph ranking model for patch weight computation. It can explore structure information among patches and unary feature of each patch together in patch weight computation. Second, we pursue a more discriminative and robust patch ranking model by further considering the temporal correlation between different frames. Third, a low-rank sparse graph is learned and incorporated in our patch ranking to build a unified learning model. A new ALM algorithm is designed to solve the unified model. Extensive experiments on two standard benchmark datasets show the effectiveness of the proposed method.

## 2 Related Work

Here, we briefly review some related works that are devoted to generate weighted patch representation for tracking. He et al. [He *et al.*, 2013] generally assumed that pixels that are far from the bounding center should be less important. He et al. [He *et al.*, 2017] proposed a key patch selection method by considering the location and occlusion. Kim et al. [Kim *et al.*, 2015] employed a random walk with restart model on a 8-neighborhood graph to compute patch weights within each bounding box. The human established structure graph may not capture the intrinsic structure among patches well and also sensitive to local noise. Li et al. [Li *et al.*, 2017a] proposed to use a subspace method to learn a more robust low-rank sparse graph for weighted patch representation. Li et al. [Li *et al.*, 2017b] also provided an improved graph learning model by considering both local and global cues together. However, these methods fail to explore the patch features and temporal correlation among different frames.

## 3 The Proposed Representation Model

Given one bounding box of the target object, we partition it into non-overlapping patches and aim to assign each patch with a foreground weight. The computed weights are combined with patch features to construct a robust object feature representation for visual tracking.

### 3.1 Patch weight learning via graph ranking

Given one bounding box in current frame, we first partition it into $n$ non-overlapping patches and aim to assign each patch with a foreground weight $\mathbf{v}_i$ to represent its possibility of belonging to the target object. We conduct this task via graph based ranking problems [Zhou *et al.*, 2003; Nie *et al.*, 2010] which first obtain several patches on target object as queries and then aim to determine the weights of rest patches according to their relevances to these queries.

Let $\mathbf{X} = (\mathbf{x}_1 \cdots \mathbf{x}_n) \in \mathbb{R}^{p \times n}$ be the collection of patch features in current $t$ frame, where $\mathbf{x}_i \in \mathbb{R}^p$ denotes the feature descriptor of patch $i$. Let $\mathbf{y} = (\mathbf{y}_1 \cdots \mathbf{y}_n)$ be the indication vector of queries, in which $\mathbf{y}_i = 1$ if patch $i$ belongs to target object, and $\mathbf{y}_i = 0$ otherwise (the detail obtaining of queries is presented in Sec. 4). Next, we define a graph $G(V, E)$ whose nodes $V$ represent patches and edges $E$ denote the spatial relationship among patches. Let $\mathbf{S}$ be the edge weight matrix. Based on features $\mathbf{X}$, graph $\mathbf{S}$ and queries $\mathbf{y}$, we compute foreground weights $\mathbf{v}$ of patches by solving,

$$\min_{\mathbf{v},\mathbf{w},b} \frac{1}{2} \sum_{i,j} \mathbf{S}_{ij}(\mathbf{v}_i - \mathbf{v}_j)^2 + \delta \|\mathbf{X}^T\mathbf{w} + \mathbf{1}b - \mathbf{v}\|_2^2 + \lambda \|\mathbf{v} - \mathbf{y}\|_2^2 \quad (1)$$

where $\mathbf{w} \in \mathbb{R}^{p \times 1}$ and $b$ is a scalar. $\mathbf{1} = (1 \cdots 1)^T \in \mathbb{R}^{n \times 1}$. The parameters $\delta, \lambda$ control the balances of the linear prediction and label fitting terms. The optimal $\mathbf{v}_i$ provides a kind of ranking for patch $i$ w.r.t queries. The ranking of patches are obtained by conducting both label prediction via linear mapping and label propagation via graph regularization. We employ this ranking as foreground weight to represent its possibility of belonging to the target object. Comparing with previous models [Kim *et al.*, 2015; Li *et al.*, 2017a] which only use structure information, this model explores both feature and structure information of patches simultaneously. To further enhance the discriminative ability and robustness of ranking, in the following we propose a more general graph optimized ranking model by further considering **temporal coherence** and **graph optimization** in ranking process.

### 3.2 Temporal coherence

The above model learns the foreground weights $\mathbf{v}$ of bounding box in current $t$ frame separately which obviously ignores the temporal correlation between current frame and previous frames. When the patch features $\mathbf{X}$ in current frame are partially contaminated or corrupted, the above separate learning model (Eq.(1)) may be less effective. To overcome this problem, we propose a temporal coherence ranking model by considering the consensus between rankings of consecutive frames. Our temporal coherence ranking model is formulated as,

$$\min_{\mathbf{v},\mathbf{w},b} \frac{1}{2} \sum_{i,j} \mathbf{S}_{ij}(\mathbf{v}_i - \mathbf{v}_j)^2 + \delta \|\mathbf{X}^T\mathbf{w} + \mathbf{1}b - \mathbf{v}\|_2^2 + \lambda \|\mathbf{v} - \mathbf{y}\|_2^2$$
$$+ \sum_{k<t} \delta_k \|\mathbf{X}^{k^T}\mathbf{w} + \mathbf{1}b - \mathbf{v}^k\|_2^2 \quad (2)$$

where $\mathbf{X}^k, \mathbf{v}^k$ denote the features and learned weights of patches in previous $k$ frame. The parameter $\delta_k$ denotes the weight of previous $k$ frame, which is determined by the correlation between objects in previous $k$ and current $t$ frames.

In above coherent ranking model, the previous features $\mathbf{X}^k$ and already learned patch rankings (weights) $\mathbf{v}^k$ are employed to guide the patch ranking $\mathbf{v}$ in current frame via the common $\mathbf{w}$ and $b$. The benefits of this coherent model are two aspects. First, when the current features $\mathbf{X}$ are partially contaminated, the current learning term $\|\mathbf{X}^T\mathbf{w} + \mathbf{1}b - \mathbf{v}\|_2^2$ may be less effective. However, the regularization terms of previous frames $\|\mathbf{X}^{k^T}\mathbf{w} + \mathbf{1}b - \mathbf{v}^k\|_2^2$ can facilitate compensating the ineffective learning of current frame and thus help to improve the robustness of the model. Second, by incorporating the patch samples of previous frames, there exist some more patch samples to learn the prediction parameters $\mathbf{w}$ and $b$, which can enhance the discriminative ability of the model.

### 3.3 Graph optimization and unified model

One important aspect of the above ranking model (Eq.(2)) is the construction of graph $\mathbf{S}$. One simple way is to use human established neighborhood graph [Kim *et al.*, 2015], which uses fixed parameters to determine the graph structure and thus usually sensitive to local noise. Inspired by graph learning work [Zhuang *et al.*, 2012], we propose to learn a graph $\mathbf{Z}$ to better capture the intrinsic relationship among patches in current frame by considering sparse, low-rank and neighborhood constraints together. By incorporating the graph learning into our coherent ranking model (Eq.(2)), we propose our final unified learning and ranking model as,

$$\min_{\mathbf{Z},\mathbf{E},\mathbf{v},\mathbf{w},b} \|\mathbf{E}\|_{2,0} + \alpha\|\mathbf{Z}\|_0 + \gamma\text{rank}(\mathbf{Z}) + \beta_1\|\mathbf{Z} - \mathbf{S}\|_2^2$$
$$+ \frac{\beta_2}{2} \sum_{i,j} \mathbf{Z}_{ij}(\mathbf{v}_i - \mathbf{v}_j)^2 + \delta\|\mathbf{X}^T\mathbf{w} + \mathbf{1}b - \mathbf{v}\|_2^2 + \lambda\|\mathbf{v} - \mathbf{y}\|_2^2$$
$$+ \sum_{k<t} \delta_k\|\mathbf{X}^{k^T}\mathbf{w} + \mathbf{1}b - \mathbf{v}^k\|_2^2$$
$$s.t. \ \mathbf{X} = \mathbf{X}\mathbf{Z} + \mathbf{E}, \mathbf{Z} \geq 0, \quad (3)$$

where $\alpha, \gamma, \beta_1, \beta_2$ are the parameters for graph learning. $\mathbf{S}$ denotes the 8-neighborhood graph, i.e., each patch is connected to 8 neighbor patches and the edge weights are computed as Gaussian kernel [Kim *et al.*, 2015]. The forth term in this model makes the final learned graph $\mathbf{Z}$ preserve the neighborhood structure of patches. In this model, the graph construction $\mathbf{Z}$ and weight computation $\mathbf{v}$ are conducted corporately for boosting their respective performance via the minimization of the fifth term, i.e., large graph affinity $\mathbf{Z}_{ij}$ encourages closeness between estimated $\mathbf{v}_i$ and $\mathbf{v}_j$ while large variation between $\mathbf{v}_i$ and $\mathbf{v}_j$ leads to lower affinity $\mathbf{Z}_{ij}$.

**Comparisons with Related Works.** Graph learning and ranking based patch weight computations have also been proposed in work DGT [Li *et al.*, 2017a] and ReGLe [Li *et al.*, 2017b]. The main differences are folows. (1) Our model employs label (weight) propagation and prediction together by exploring both structure and feature information of patches to enhance the discriminative ability of learning model. In contrast, previous works only employ label propagation with structure information and ignore the feature information although the feature information has been used in their graph constructions. (2) Our model learns the patch weight by exploring both spatial consistency and temporal coherence simultaneously while previous works conduct patch weight computation in each frame separately and only exploit spatial consistency.

## 4 Optimization

It is known that both $rank$ function and $\ell_0$ norm are non-convexity. Thus, it is difficult to optimize problem Eq.(3) directly. One popular and effective way is to replace $rank$ function and $\ell_0$ norm with nuclear norm and $\ell_1$ norm and reformulate Eq.(3) approximately as the following convex problem,

$$\min_{\mathbf{Z,E,v,w},b} \|\mathbf{E}\|_{2,1} + \alpha\|\mathbf{Z}\|_1 + \gamma\|\mathbf{Z}\|_* + \beta_1\|\mathbf{Z-S}\|_2^2$$
$$+ \tfrac{\beta_2}{2}\sum_{i,j}\mathbf{Z}_{ij}(\mathbf{v}_i-\mathbf{v}_j)^2 + \delta\|\mathbf{X}^T\mathbf{w}+\mathbf{1}b-\mathbf{v}\|_2^2 + \lambda\|\mathbf{v-y}\|_2^2$$
$$+ \sum_{k<t}\delta_k\|\mathbf{X}^{k^T}\mathbf{w}+\mathbf{1}b-\mathbf{v}^k\|_2^2$$
$$s.t.\ \mathbf{X=XZ+E, Z}\geq 0, \tag{4}$$

Here $\|\mathbf{Z}\|_*, \|\mathbf{Z}\|_1=\sum_{i,j}|\mathbf{Z}_{ij}|$ and $\|\mathbf{E}\|_{2,1}=\sum_i\sqrt{\sum_j \mathbf{E}_{ij}^2}$ denote the nuclear norm, $\ell_1$ norm and $\ell_{2,1}$ norm, respectively. This problem is a convex problem and the global optimal solution can be obtained via an Augmented Lagrange Multiplier Method (ALM) [Lin *et al.*, 2011] algorithm. To do so, we first introduce auxiliary variables $\mathbf{A, B, U}$ to make the objective function separable as,

$$\min_{\mathbf{Z,E,v,w},b} \|\mathbf{E}\|_{2,1} + \alpha\|\mathbf{U}\|_1 + \gamma\|\mathbf{B}\|_* + \beta_1\|\mathbf{Z-S}\|_2^2$$
$$+ \tfrac{\beta_2}{2}\sum_{i,j}\mathbf{A}_{ij}(\mathbf{v}_i-\mathbf{v}_j)^2 + \delta\|\mathbf{X}^T\mathbf{w}+\mathbf{1}b-\mathbf{v}\|_2^2 + \lambda\|\mathbf{v-y}\|_2^2$$
$$+ \sum_{k<t}\delta_k\|\mathbf{X}^{k^T}\mathbf{w}+\mathbf{1}b-\mathbf{v}^k\|_2^2$$
$$s.t.\ \mathbf{X=XZ+E, Z=U, Z=B, Z=A, A}\geq 0, \tag{5}$$

The augmented Lagrangian function $\mathcal{L}(\mathbf{Z,E,A,B,U})$ is,

$$\mathcal{L} = \|\mathbf{E}\|_{2,1} + \alpha\|\mathbf{U}\|_1 + \gamma\|\mathbf{B}\|_* + \beta_1\|\mathbf{Z-S}\|^2$$
$$+ \tfrac{\beta_2}{2}\sum_{i,j}\mathbf{A}_{ij}(\mathbf{v}_i-\mathbf{v}_j)^2 + \langle\mathbf{Y}_1,\mathbf{E-X+XZ}\rangle$$
$$+ \langle\mathbf{Y}_2,\mathbf{Z-U}\rangle + \langle\mathbf{Y}_3,\mathbf{Z-B}\rangle + \langle\mathbf{Y}_4,\mathbf{Z-A}\rangle \tag{6}$$
$$+ \tfrac{\mu}{2}(\|\mathbf{X-XZ-E}\|^2 + \|\mathbf{Z-U}\|^2 + \|\mathbf{Z-B}\|^2 + \|\mathbf{Z-A}\|^2)$$
$$+ \delta\|\mathbf{X}^T\mathbf{w}+\mathbf{1}b-\mathbf{v}\|_2^2 + \lambda\|\mathbf{v-y}\|_2^2 + \sum_{k<t}\delta_k\|\mathbf{X}^{k^T}\mathbf{w}+\mathbf{1}b-\mathbf{v}^k\|_2^2$$

$$= \|\mathbf{E}\|_{2,1} + \alpha\|\mathbf{U}\|_1 + \gamma\|\mathbf{B}\|_* + \beta_1\|\mathbf{Z-S}\|^2$$
$$+ \tfrac{\beta_2}{2}\sum_{i,j}\mathbf{A}_{ij}(\mathbf{v}_i-\mathbf{v}_j)^2 + \mathcal{H}(\mathbf{Z,E,B,U,A},\mathbf{Y}_1\sim\mathbf{Y}_4)$$
$$- \tfrac{1}{2\mu}(\|\mathbf{Y}_1\|^2+\|\mathbf{Y}_2\|^2+\|\mathbf{Y}_3\|^2+\|\mathbf{Y}_4\|^2) \tag{7}$$
$$+ \delta\|\mathbf{X}^T\mathbf{w}+\mathbf{1}b-\mathbf{v}\|_2^2 + \lambda\|\mathbf{v-y}\|_2^2 + \sum_{k<t}\delta_k\|\mathbf{X}^{k^T}\mathbf{w}+\mathbf{1}b-\mathbf{v}^k\|_2^2$$

where

$$\mathcal{H}(\mathbf{Z,E,B,U,A}) = \tfrac{\mu}{2}(\|\mathbf{X-XZ-E}+\tfrac{\mathbf{Y}_1}{\mu}\|^2$$
$$+ \|\mathbf{Z-U}+\tfrac{\mathbf{Y}_2}{\mu}\|^2 + \|\mathbf{Z-B}+\tfrac{\mathbf{Y}_3}{\mu}\|^2 + \|\mathbf{Z-A}+\tfrac{\mathbf{Y}_4}{\mu}\|^2$$

The ALM algorithm updates the variables $\mathbf{Z,E,U,B,A,w},b$ and $\mathbf{v}$ alternatively. Let $\mathbf{R}\in\mathbb{R}^{n\times n}$ and $\mathbf{R}_{ij}=(\mathbf{v}_i-\mathbf{v}_j)^2$. Then, with some algebra, the update rules for these variables are derived as follows,

$$\mathbf{U} = \arg\min_{\mathbf{U}} \alpha\|\mathbf{U}\|_1 + \tfrac{\mu}{2}\|\mathbf{Z-U}+\tfrac{\mathbf{Y}_2}{\mu}\|^2$$
$$= S_{\alpha\mu^{-1}}(\mathbf{Z}+\tfrac{\mathbf{Y}_2}{\mu}) \tag{8}$$

$$\mathbf{B} = \arg\min_{\mathbf{B}} \gamma\|\mathbf{B}\|_* + \tfrac{\mu}{2}\|\mathbf{Z-B}+\tfrac{\mathbf{Y}_3}{\mu}\|^2$$
$$= \Theta_{\gamma\mu^{-1}}(\mathbf{X-XZ}+\tfrac{\mathbf{Y}_1}{\mu}) \tag{9}$$

$$\mathbf{A} = \arg\min_{\mathbf{A}\geq 0} \tfrac{\beta_2}{2}\sum_{i,j}\mathbf{A}_{ij}(\mathbf{v}_i^t-\mathbf{v}_j^t)^2 + \tfrac{\mu}{2}\|\mathbf{Z-A}+\tfrac{\mathbf{Y}_4}{\mu}\|^2$$
$$= \left[\mathbf{Z}+\tfrac{\mathbf{Y}_4}{\mu}-\tfrac{\beta_1}{\mu}\mathbf{R}\right]_+ \tag{10}$$

$$\mathbf{E} = \arg\min_{\mathbf{E}} \|\mathbf{E}\|_{2,1} + \tfrac{\mu}{2}\|\mathbf{X}^t-\mathbf{X}^t\mathbf{Z-E}+\tfrac{\mathbf{Y}_1}{\mu}\|^2$$
$$= \Omega_{\mu^{-1}}(\mathbf{X-XZ}+\tfrac{\mathbf{Y}_1}{\mu}) \tag{11}$$

$$\mathbf{Z} = \arg\min_{\mathbf{Z}} \beta_1\|\mathbf{Z-S}\|^2 + \mathcal{H}(\mathbf{Z,E,B,U,A}) \tag{12}$$

$$\mathbf{v} = \arg\min_{\mathbf{v}} \tfrac{\beta_2}{2}\sum_{i,j}\mathbf{Z}_{ij}(\mathbf{v}_i-\mathbf{v}_j)^2 + \delta\|\mathbf{X}^T\mathbf{w}+\mathbf{1}b-\mathbf{v}\|_2^2$$
$$+ \lambda\|\mathbf{v-y}\|_2^2$$
$$= ((\delta_t+\lambda)\mathbf{I}+\beta_2(\mathbf{D-Z}))^{-1}(\lambda\mathbf{y}+\delta_t(\mathbf{X}^T\mathbf{w}+\mathbf{1}b)) \tag{13}$$

$$\mathbf{w} = \arg\min_{\mathbf{w},b} \delta\|\mathbf{X}^T\mathbf{w}+\mathbf{1}b-\mathbf{v}\|_2^2 + \sum_{k<t}\delta_k\|\mathbf{X}^{k^T}\mathbf{w}+\mathbf{1}b-\mathbf{v}^k\|_2^2$$
$$= \left(\sum_{k<t}\delta_k\mathbf{X}^k\mathbf{X}^{k^T}+\delta\mathbf{X}\mathbf{X}^T\right)^{-1}\left[\sum_{k<t}\delta_k\mathbf{X}^k(\mathbf{1}b-\mathbf{v}^k)+\delta\mathbf{X}(\mathbf{1}b-\mathbf{v})\right] \tag{14}$$

$$b = \arg\min_{\mathbf{w},b} \delta\|\mathbf{X}^T\mathbf{w}+\mathbf{1}b-\mathbf{v}\|_2^2 + \sum_{k<t}\delta_k\|\mathbf{X}^{k^T}\mathbf{w}+\mathbf{1}b-\mathbf{v}^k\|_2^2$$
$$= -\frac{\sum_{k<t}\delta_k\mathbf{1}^T(\mathbf{X}^{k^T}\mathbf{w}-\mathbf{v}^k)+\delta\mathbf{1}^T(\mathbf{X}^T\mathbf{w}-\mathbf{v})}{\sum_{k<t}\delta_k+\delta} \tag{15}$$

where $\Omega, S, \Theta$ are the $\ell_{2,1}$ minimization, shrinkage and singular value thresholding operator, respectively. In Eq.(13), $\mathbf{D}$ is a diagonal matrix with $\mathbf{D}_{ii}=\sum_j\mathbf{Z}_{ij}$. Note that, the update

$\mathbf{Z}$ in Eq.(12) has a closed-form solution and can be obtained by setting the first derivation w.r.t $\mathbf{Z}$ to zero. The closed-form solution of $\mathbf{Z}$ is

$$\mathbf{Z} = \left(\mathbf{X}^T\mathbf{X} + 3\mathbf{I} + \tfrac{2\beta_1}{\mu}\mathbf{I}\right)^{-1}\left[\mathbf{X}^T\mathbf{X} - \mathbf{X}^T\mathbf{E} + \mathbf{B}\right.$$
$$\left. + \mathbf{A} + \mathbf{U} + \tfrac{2\beta_1}{\mu}\mathbf{S} + \tfrac{1}{\mu}(\mathbf{X}^T\mathbf{Y}_1 - \mathbf{Y}_2 - \mathbf{Y}_3 - \mathbf{Y}_4)\right] \quad (16)$$

The updates $\mathbf{v}, \mathbf{w}$ and $b$ in Eqs.(13,14,15) also have closed-form solutions and are obtained by setting the first derivation w.r.t variable $\mathbf{v}, \mathbf{w}$ and $b$ to zero, respectively. The complete algorithm is summarized in Algorithm 1.

---

**Algorithm 1** Optimization of Problem Eq.(4)

---

**Input:** The patch features $\mathbf{X}^k, k \leq t$. The previous learned patch weights $\mathbf{v}^k, k < t$, the query indication vector $\mathbf{y}$, the human constructed graph $\mathbf{S}$.
Set $\mathbf{E} = \mathbf{Z} = \mathbf{U} = \mathbf{B} = \mathbf{A} = \mathbf{0}$, $\mathbf{Y}_1 = \mathbf{Y}_2 = \mathbf{Y}_3 = \mathbf{Y}_4 = \mathbf{0}$, $\rho = 1.3$, $\mu_{max} = 10^{10}, \epsilon = 10^{-6}, \mu_0 = 10^{-6}$, maxIter $= 50$ and $k = 0$.
**Output:** Patch weights $\mathbf{v}$ of current frame.
1: **while** not converged **do**
2:     Update $\mathbf{U}, \mathbf{B}, \mathbf{A}, \mathbf{E}$ by Eq.(8-11), respectively;
3:     Update $\mathbf{Z}$ by Eq.(16);
4:     Update $\mathbf{v}$ by Eq.(13);
5:     Update $\mathbf{w}$ by Eq.(14);
6:     Update $b$ by Eq.(15);
7:     Update Lagrangian multipliers as follows,
8:         $\mathbf{Y}_1 = \mathbf{Y}_1 + \mu(\mathbf{X} - \mathbf{XZ} - \mathbf{E})$
9:         $\mathbf{Y}_2 = \mathbf{Y}_2 + \mu(\mathbf{Z} - \mathbf{U})$
10:       $\mathbf{Y}_3 = \mathbf{Y}_3 + \mu(\mathbf{Z} - \mathbf{B})$
11:       $\mathbf{Y}_4 = \mathbf{Y}_4 + \mu(\mathbf{Z} - \mathbf{A})$
12:    Update $\mu$ by $\mu = \min(\mu_{max}, \rho\mu)$;
13:    Update $k$ by $k = k + 1$;
14:    Check the convergence condition, i.e, the maximum element changes of $\mathbf{E}, \mathbf{Z}, \mathbf{U}, \mathbf{B}, \mathbf{A}, \mathbf{v}, \mathbf{w}$ and $b$ between two consecutive iterations are less than $\epsilon$ or the maximum number of iterations reaches maxIter.
15: **end while**

---

## 5 Visual Object Tracking

In this section, we incorporate our optimized weights of patches into the tracking-by-detection framework, Struck [Hare et al., 2011] to provide a robust tracking algorithm. Our tracking process contains two main steps: weighted patch feature descriptor and structured output tracking.

**Weighted patch feature descriptor**. Similar to [Kim et al., 2015; Li et al., 2017a], we obtain the foreground queries as follows. For each patch in the bounding box, if it belongs to the shrunk region of the bounding box then we regard it as foreground query because it is very likely to belong to target object, as shown in Figure 1 (b). Based on these foreground queries, we can thus obtain the foreground weights $\mathbf{v}$ for patches using the proposed ranking model, i.e., larger weight $\mathbf{v}_i$ indicates the more likely that patch $i$ belongs to target. In addition to foreground queries, we can also obtain background queries by using patches in the expended region of bounding box (as shown in Figure 1 (c)) and obtain background weights $\mathbf{u}$ for patches similarly using the proposed ranking model, i.e., larger weight $\mathbf{u}_i$ indicates the more likely that patch $i$ belongs to background. We can thus combine $\mathbf{v}$ and $\mathbf{u}$ together as $w_i = \frac{1}{1+\exp(-\epsilon(\mathbf{v}_i - \mathbf{u}_i))}, \epsilon = 43$ to provide a more accurate foreground weights [Kim et al., 2015]. Based on these patch weights, we can thus generate a kind of feature representation for visual object by incorporating patch weights into feature descriptor. Formally, let $\mathbf{X}^c = (\mathbf{x}_1^c, \mathbf{x}_2^c \cdots \mathbf{x}_n^c)$ be the original feature descriptor for bounding box $c$, where $\mathbf{x}_i^c$ denotes the feature of patch $i$. By incorporating the learned weights into the feature descriptor $\mathbf{X}^c$, we can thus obtain a kind of weighted descriptor for the bounding box as

$$\mathbf{X}_w^c = (w_1\mathbf{x}_1^c, w_2\mathbf{x}_2^c \cdots w_n\mathbf{x}_n^c) \quad (17)$$

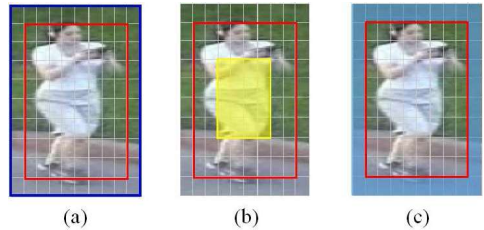

Figure 1: (a) Original bounding box (red box) and expended bounding box (blue box); (b) Shrunk region (yellow region); (c) Expended region (blue region).

Comparing with original feature $\mathbf{X}^c$, the weighted feature descriptor $\mathbf{X}_w^c$ can alleviate the effects of cluttered background information and thus provides a kind of more accurate feature descriptor for target in tracking process [Kim et al., 2015].

**Structured SVM tracking**. We apply our weighted patch descriptor $\mathbf{X}_w$ into Struck tracking algorithm, which aims to determine the optimal object bounding box in the current frame by maximizing the classification score,

$$c^* = \arg\max_c \langle \mathbf{h}_{t-1}, \mathbf{X}_w^c \rangle \quad (18)$$

where $\mathbf{h}_{t-1}$ is the normal vector of a decision plan of $(t-1)$th frame. In order to further incorporate the information of the initial frame and previous frame, in this paper we compute the optimal bounding box by maximizing the following score,

$$c^* = \arg\max_c \left(\alpha_1 \langle \mathbf{h}_{t-1}, \mathbf{X}_w^c \rangle + \alpha_2 \langle \mathbf{h}_{t-2}, \mathbf{X}_w^c \rangle + \alpha_3 \langle \mathbf{h}_1, \mathbf{X}_w^c \rangle\right),$$
$$(19)$$

where $\mathbf{h}_1, \mathbf{h}_{t-2}$ is learned in first frame and previous $t-2$ frame, respectively. This strategy can prevent it from learning drastic appearance changes. When the optimal bounding box $c^*$ is estimated, we then update the current classifier $\mathbf{h}_t$. To prevent the effects of unreliable tracking results, here we update the classifier only when the confidence score of tracking result is larger than a threshold $\theta = 0.3$ [Kim et al., 2015].

## 6 Experiments

We evaluate the effectiveness of the proposed tracking algorithm on two widely used benchmark datasets [Liang et al., 2015; Wu et al., 2015] and compare with some other state-of-the-art trackers. We implement our tracker using C++ language on a desktop computer with an Inter i7 3.6GHz CPU and 32GB RAM. The proposed tracker performs at about 3.3 FPS (frames per second). In general, our tracker performs slower than DGT [Li *et al.*, 2017a], but faster than ReGLe [Li *et al.*, 2017b] in which 2 FPS has been reported.

Table 1: Comparison of attribute-based PR/SR scores on OTB benchmark dataset. The attributes includes IV (illumination variation), SV (scale variation), OCC (occlusion), DEF (deformation), MB(motion blur), FM (fast motion), IPR (in-plane-rotation), OPR (out-of-plane rotation), OV (out-of-view), BC (background clutters), and LR (low resolution). The best, second best and third best performances are indicated by red, green and blue colors.

|     | Staple | MEEM | SOWP | ReGLe | ACFN | LCT | SRDCF | HCF | DGT | Ours |
| --- | --- | --- | --- | --- | --- | --- | --- | --- | --- | --- |
| FM  | 0.670/0.501 | 0.752/0.542 | 0.723/0.556 | 0.802/0.588 | 0.758/0.566 | 0.681/0.534 | 0.768/0.598 | 0.814/0.570 | 0.777/0.549 | 0.790/0.608 |
| BC  | 0.716/0.524 | 0.746/0.519 | 0.775/0.570 | 0.841/0.612 | 0.769/0.542 | 0.734/0.550 | 0.774/0.584 | 0.842/0.585 | 0.867/0.614 | 0.840/0.619 |
| MB  | 0.642/0.493 | 0.731/0.556 | 0.702/0.567 | 0.791/0.601 | 0.731/0.568 | 0.669/0.533 | 0.766/0.595 | 0.803/0.585 | 0.815/0.591 | 0.786/0.607 |
| DEF | 0.712/0.514 | 0.754/0.489 | 0.741/0.527 | 0.858/0.580 | 0.772/0.535 | 0.689/0.499 | 0.734/0.544 | 0.778/0.523 | 0.857/0.582 | 0.851/0.590 |
| IV  | 0.772/0.551 | 0.728/0.515 | 0.766/0.554 | 0.837/0.593 | 0.777/0.554 | 0.732/0.557 | 0.781/0.601 | 0.792/0.532 | 0.838/0.573 | 0.838/0.592 |
| IPR | 0.756/0.520 | 0.794/0.529 | 0.828/0.567 | 0.848/0.579 | 0.785/0.546 | 0.782/0.557 | 0.744/0.544 | 0.853/0.559 | 0.856/0.573 | 0.828/0.578 |
| LR  | 0.773/0.406 | 0.808/0.382 | 0.903/0.423 | 0.936/0.514 | 0.818/0.515 | 0.699/0.399 | 0.765/0.514 | 0.847/0.388 | 0.732/0.417 | 0.908/0.511 |
| OCC | 0.715/0.520 | 0.741/0.504 | 0.754/0.528 | 0.836/0.585 | 0.756/0.542 | 0.682/0.507 | 0.735/0.560 | 0.755/0.520 | 0.820/0.562 | 0.824/0.574 |
| OPR | 0.734/0.523 | 0.794/0.525 | 0.787/0.547 | 0.847/0.576 | 0.777/0.543 | 0.746/0.538 | 0.742/0.550 | 0.807/0.534 | 0.855/0.577 | 0.843/0.586 |
| OV  | 0.594/0.446 | 0.685/0.488 | 0.633/0.497 | 0.794/0.565 | 0.692/0.508 | 0.592/0.452 | 0.596/0.463 | 0.676/0.474 | 0.753/0.533 | 0.755/0.553 |
| SV  | 0.732/0.506 | 0.736/0.470 | 0.746/0.475 | 0.825/0.552 | 0.764/0.551 | 0.681/0.488 | 0.745/0.562 | 0.790/0.481 | 0.813/0.504 | 0.820/0.549 |
| ALL | 0.755/0.537 | 0.781/0.530 | 0.803/0.560 | 0.869/0.608 | 0.802/0.575 | 0.762/0.562 | 0.789/0.598 | 0.831/0.559 | 0.865/0.586 | 0.866/0.612 |

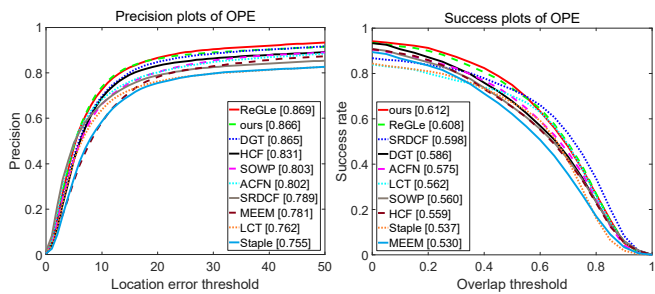

Figure 2: Precision plots and success plots of OPE (one-pass evaluation) of the proposed tracker against other state-of-the-art trackers on OTB100 dataset.

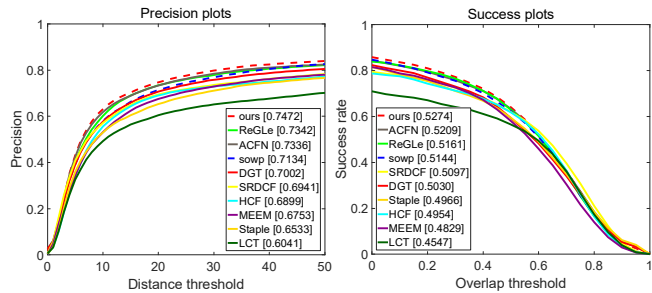

Figure 3: Evaluation results on TColor-128 dataset, The legend contains representative PR and SR values. Note that, our method performs favorably against the state-of-the-art trackers.

### 6.1 Evaluation settings

**Parameters**. For each bounding box in each frame, we partition it into $8 \times 8$ no-overlapping patches to obtain better performance. For each patch, we extract a 32-dimensional feature descriptor including a 24-dimensional RGB color histogram and 8-dimensional oriented gradient histogram. For efficiency consideration, we scale each frame and set the minimum side length of bounding box to 32 pixels. We set the side length of a searching window as $2\sqrt{wh}$, where $w$ and $h$ denote the width and height of the scaled bounding box, respectively. We set parameters $\{\alpha, \gamma, \beta_1, \beta_2, \lambda\}$ to $\{0.2, 0.08, 20, 0.9, 1.0\}$. Empirically, the tracking performance is insensitive to these parameters. For efficiency consideration, we use previous $(t-1)$ frame and first initial frame in our model and set balance parameters $\{\delta, \delta_{t-1}, \delta_1\}$ to $\{0.3, 0.3, 0.3\}$. Because the previous $(t-1)$ frame is most correlated with current $t$ frame and the first initial frame provides a ground truth bounding box. We set the balanced parameters in Eq.(19) $\{\alpha_1, \alpha_2, \alpha_3\}$ to $\{0.63, 0.07, 0.03\}$

**OTB100 benchmark dataset**. This dataset [Wu *et al.*, 2015] contains 100 image sequences whose ground-truth target locations are marked manually. The sequences are associated with 11 different attributes. For performance evaluation metrics, we use precision rate (PR) and success rate (SR) [Henriques *et al.*, 2015; Wu *et al.*, 2015] to measure the quantitative performances of tracker methods.

**Temple-Color benchmark dataset**. This dataset [Liang *et al.*, 2015] contains 128 challenging image sequences of human, animals and rigid objects, whose ground-truth target locations are marked manually. Each sequence in this dataset is also annotated by its challenge factors, which is the same as in [Wu *et al.*, 2015]. The evaluation metrics including PR and SR used in this dataset are also same with [Wu *et al.*, 2015].

### 6.2 Evaluation on OTB100 dataset

We first present the evaluation results on OTB 100 dataset and compare our method with some other state-of-the-art methods including both deep and non-deep learning methods.

**Comparison with non-deep learning trackers**. We compare our tracking method with some recent state-of-the-art traditional methods including DGT [Li *et al.*, 2017a], ReGLe [Li *et al.*, 2017b], SOWP [Kim *et al.*, 2015], Staple [Bertinetto *et al.*, 2016], SRDCF [Danelljan *et al.*, 2015], MEEM [Zhang *et al.*, 2014] and LCT [Ma *et al.*, 2015a]. Note that, DGT [Li *et al.*, 2017a], ReGLe [Li *et al.*, 2017b] and SOWP [Kim *et al.*, 2015] also use the weighted patch representation in tracking process and thus are most related with our methods. Figure 2 shows the comparison results in one-pass evaluation (OPE) using PR and SR curves, respectively. Overall, our tracking method generally performs better than the other state-of-the-art methods. In particular, our method achieves 0.1%/2.6%, 6.3%/5.2% performance gains in PR/SR over related work DGT [Li *et al.*, 2017a] and SOWP [Kim *et al.*, 2015] and 0.4% performance gains in SR over ReGLe [Li *et al.*, 2017b], which demonstrates the effectiveness of the proposed weighted patch representation model in conducting object tracking task. Note that, comparing with other related works, our model further explores the feature information via label prediction and also exploits temporal correlation in visual object representation and thus performs more discriminatively and robustly.

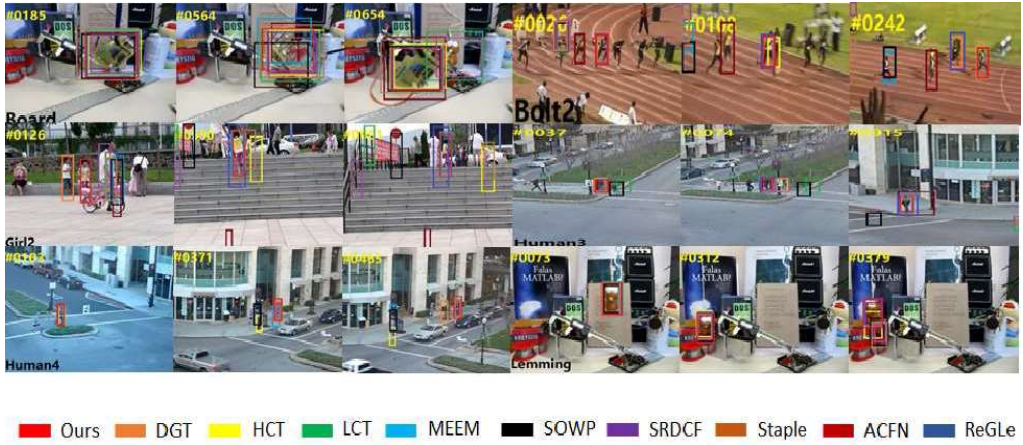

Figure 4: Tracking results of our method against 8 trackers (denoted in different colors and lines) on six challenging sequences. Intuitively, one can note that our tracker locates the visual object more accurately on these challenging sequences.

**Comparison with deep learning trackers**. In Figure 2, we also report some results of the deep learning based tracking methods including ACFN [Choi *et al.*, 2017] and HCF [Ma *et al.*, 2015b]. Our method obtains better performance than these deep learning trackers. Note that, deep learning based tracking methods generally require large-scale annotated training samples while our method only uses the ground truth annotation in the first frame to train our model and then updates the model in subsequent frames.

**Evaluation on different attributes**. We present the representative PR/SR values on videos belonging to 11 different attributes, respectively. Table 1 reports the comparison results (PR/SR) on sequences that belong to 11 different attributes, respectively. One can note that, our method obtains the best performance on most challenging attributes in SR. Figure 4 shows some tracking examples on some challenging videos. Intuitively, one can note that our tracker locates the visual object more accurately on these challenging sequences.

### 6.3 Evaluation on Temple-Color dataset

We also evaluate our method on Temple-Color dataset [Liang *et al.*, 2015]. Figure 3 shows the success plot and precision plot over all 129 videos on this dataset. Generally, our tracker outperforms the other related trackers and obtains the best performance on PR/SR values. Especially, it achieves 3.38%/1.3%, 4.7%/2.44% and 1.3%/1.13% performance gains in PR/SR over most related work SOWP, DGT and ReGLe. This further demonstrates the effectiveness of the proposed tracking method.

### 6.4 Component analysis

To justify the importance of two main components (temporal coherence and graph learning) in our representation model, we implement some special variants of our model, i.e., Ours-noT and Ours-noG. 1) Ours-noT only uses the information of current frame and does not exploit the temporal correlation among frames in our model. 2) Ours-noG only uses the human established graph **S** [Kim *et al.*, 2015] and does not exploit the graph learning **Z** in our model. Figure 5 shows the SR scores on videos of 11 different attributes. We can note that, (1) Utilizing the graph learning is obviously beneficial in our weighted patch representation and thus tracking performance. (2) The temporal relationship among patches is an important cue to obtain robust weighted patch representation.

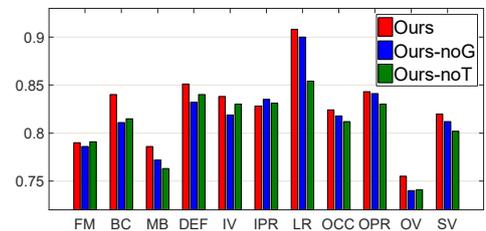

Figure 5: Performance of two variants of the proposed model on videos of 11 different attributes on OTB100 dataset.

## 7 Conclusion

This paper proposes a novel unified temporal coherence and graph optimized ranking model for weighted patch object representation and visual tracking problem. The proposed unified model integrates the cues of spatial structure, temporal correlation and unary features together to and thus performs more robustly and discriminatively in patch weight computation. We incorporate the optimized weighted patch representation into Struck tracker to carry out visual object tracking. Experiments on two standard benchmark datasets show the effectiveness of the proposed tracking method.